\documentclass[final, twocolumn,5p,times, authoryear]{elsarticle}
\usepackage{amsmath}
\usepackage{booktabs}
\usepackage{url}
\usepackage{algorithm}
\usepackage{algpseudocode}
\usepackage{array}
\usepackage[colorlinks=False]{hyperref}
\usepackage[round]{natbib}
\usepackage{lineno}

\usepackage{amsmath}

\usepackage{array}
\usepackage{microtype}
\usepackage{booktabs}
\usepackage{tabularx}
\usepackage{multirow}
\usepackage{amsfonts}

\usepackage{graphicx}

\usepackage{url}

\usepackage{orcidlink}

\usepackage{algorithm}
\usepackage{algpseudocode}

\usepackage{placeins}

\begin{document}
\begin{frontmatter}

\title{Test-Time Adaptation for Height Completion via Self-Supervised ViT Features and Monocular Foundation Models}

\author[inst1]{Osher Rafaeli\corref{cor1}\raisebox{0.5ex}{\orcidlink{0000-0002-7097-7568}}}
\ead{osherr@post.bgu.ac.il}
\cortext[cor1]{Corresponding author.}

\author[inst1,inst2]{Tal Svoray\raisebox{0.5ex}{\orcidlink{0000-0003-2243-8532}}}

\author[inst1]{Ariel Nahlieli\raisebox{0.5ex}{\orcidlink{0009-0001-0633-1842}}}

\affiliation[inst1]{
    organization={Department of Environmental, Geoinformatics and Urban Planning Sciences, 
    Ben-Gurion University of the Negev},
    country={Israel}
}

\affiliation[inst2]{
    organization={Department of Psychology, 
    Ben-Gurion University of the Negev},
    country={Israel}
}

\begin{abstract}
Accurate digital surface models (DSMs) are essential for many geospatial applications, including urban monitoring, environmental analyses, infrastructure management, and change detection. However, large-scale DSMs frequently contain incomplete or outdated regions due to acquisition limitations, reconstruction artifacts, or changes in the built environment. Traditional height completion approaches primarily rely on spatial interpolation or which assume spatial continuity and therefore fail when objects are missing. Recent learning-based approaches improve reconstruction quality but typically require supervised training on sensor-specific datasets, limiting their generalization across domains and sensing conditions. We propose Prior2DSM, a training-free framework for metric DSM completion that operates entirely at test time by leveraging foundation models. Unlike previous height completion approaches that require task-specific training, the proposed method combines self-supervised Vision Transformer (ViT) features from DINOv3 with monocular depth foundation models to propagate metric information from incomplete height priors through semantic feature-space correspondence. Test-time adaptation (TTA) is performed using parameter-efficient low-rank adaptation (LoRA) together with a lightweight multilayer perceptron (MLP), which predicts spatially varying scale and shift parameters to convert relative depth estimates into metric heights. Experiments conducted on high-resolution aerial (NAIP) and satellite (WorldView-3) datasets, under varying levels of prior incompleteness, demonstrate consistent improvements over interpolation based methods, prior-based rescaling height approaches, and state-of-the-art monocular depth estimation models. Prior2DSM reduces reconstruction error while preserving structural fidelity, achieving up to a 46\% reduction in RMSE compared to linear fitting of MDE, and further enables DSM updating and coupled RGB--DSM generation.
\end{abstract}

\begin{keyword}
Digital Surface Model (DSM) \sep Monocular Height Estimation \sep Vision Transformers \sep Foundation Models \sep Test-Time Adaptation \sep Remote Sensing
\end{keyword}

\end{frontmatter}


\section{Introduction}

Precise measurement and characterization of three-dimensional (3D) surface features are essential for a wide range of geoinformatics applications, including urban planning, environmental monitoring, and disaster assessment \citep{ARAVENAPELIZARI2026357}. In particular, accurate estimation of buildings, forests and terrain heights, from remotely sensed (RS) data, constitutes a critical component for Earth surface analyses and diverse scientific and engineering tasks \citep{OKOLIE20221}.

Conventional height estimation techniques, such as stereo photogrammetry and airborne LiDAR scanning, are widely used and provide reliable results across various surface types \citep{hladik2012accuracy}. However, photogrammetric techniques require extensive collections of cloud-free, high-resolution optical imagery, while LiDAR acquisition is cost-prohibitive at large spatial scales \citep{BHARDWAJ2016125, COLOMINA, frsen}. Consequently, availability of accurate and up-to-date height information, over large areas, remains limited \citep{wang2021challenges}.

Growing availability of high-resolution satellite and aerial imagery has improved access to single-view optical data \citep{10758818}, driving interest in monocular height estimation (MHE) methods that exploit 2D image features. Recent advances in convolutional neural networks (CNNs), and Vision Transformers (ViTs), coupled with large-scale foundation models success in computer vision, have driven the development of MHE techniques also in RS studies using overhead imagery \citep{SONG2026155, 10294289, hong2025depth2elevation, cambrin2024depth}. 
Although these methods have progressively improved predictive performance \citep{8128167}, MHE in RS remains an ill-posed problem: near-orthographic imaging geometry eliminates perspective height cues, resulting in severe scale-height ambiguity \citep{10294289, rs12172719}. Consequently, existing methods often exhibit limited robustness and strong dataset bias, leading to poor generalization to unseen domains \citep{zeng2024rsaresolvingscaleambiguities, hong2025depth2elevation}. Pretrained MHE models therefore struggle in demanding applications, particularly under domain shifts  \citep{10294289, essd-17-6647-2025}. Tasks such as metric change detection, void filling, and correction of photogrammetric artifacts require stronger geometric accuracy .

Recently, methods incorporating metric height priors, have gained attention, because incomplete height measurements can provide absolute metric constraints that resolve scale ambiguity by anchoring predictions to geodetic reference points \citep{SONG2026155}. However, existing non-parametric rescaling strategies operate directly in the height domain through spatial interpolation or linear alignment-based rescaling, which becomes insufficient when surface objects are completely missing \citep{lin2024promptda, guo2025multiviewreconstructionsfmguidedmonocular, zhong2025novelsolutiondronephotogrammetry}. Deep feature-domain reconstruction is indeed promising for height inpainting, by exploiting learned structural priors beyond local spatial neighborhoods \citep{chen2026reconstructingbuildingheightspaceborne, song2025enhancing}. 

Concurrently, proliferation of multi-source height data, from photogrammetric DSMs, SAR interferometry, and legacy surveys, has created widespread availability of incomplete, or outdated, height data, that could serve as metric anchors, if effectively integrated with MHE.
However, prior-based height completion approaches are typically constrained to specific prior sources and object categories (e.g., buildings), and rely heavily on supervised training, making them susceptible to dataset bias with limited robustness \citep{chen2026reconstructingbuildingheightspaceborne}. The fundamental challenge is that height discontinuities, and fully missing regions, violate spatial continuity assumptions underlying interpolation-based methods, while category-specific supervised approaches cannot generalize beyond their training distributions. This necessitates a mechanism to infer height relationships from semantically analogous regions, regardless of spatial separation, requiring explicit coupling between semantic understanding and geometric reconstruction, that current methods do not provide. They are trapped between requiring complete spatial coverage (interpolation), being sensor-specific (supervised methods), or lacking metric anchoring (pure monocular inference).

We adopt self-supervised ViT features that encode rich dense semantic representations \citep{amir2022deepvitfeaturesdense} and monocular foundation models, to estimate relative height geometry \citep{Yang2024DepthV2, bochkovskii2025depthprosharpmonocular, wang2025moge2accuratemonoculargeometry}. We reformulate height alignment as a feature-guided correction, where each pixel is adjusted based on relative height patterns observed in semantically similar regions, identified through deep feature space. This exploits the implicit coupling between semantic similarity and geometric consistency encoded in self-supervised features, enabling height transfer across spatially distant, but semantically analogous.

\begin{itemize}

\item We introduce a test time adaptation (TTA) and training-independent height completion framework that exploits self-supervised feature-space similarity, for robust local metric alignment. Unlike spatial interpolation or global affine rescaling, our method propagates height information through embedding-guided  regression, enabling completion of fully missing objects under varying prior incompleteness levels.

\item We demonstrate consistent and substantial improvements in metric accuracy (MAE/RMSE), and structural fidelity (SSIM), over pure monocular reconstruction models, and state-of-the-art prior-based depth completion methods, with particularly pronounced gains in complex urban and high-relief environments.

\item We systematically benchmark state-of-the-art monocular foundation backbones (Depth Anything V2, Depth Pro, and MoGe-2), analyzing their impact on completion accuracy, domain robustness, and sensitivity to prior incompleteness.

\item We validate the practical applicability of Prior2DSM through real-world DSM updating (2008–present temporal correction) and RGB–DSM co-editing via integration with a remote sensing text-to-image generation model, demonstrating robust metric reconstruction under both real and synthetic scenarios.

\end{itemize}
\begin{table*}[t]
\centering
\caption{Comparison of representative height completion approaches. 
\checkmark~indicates full support, $\times$ indicates no support, and $\circ$ indicates limited support.}
\setlength{\tabcolsep}{2pt}
\label{tab:comparison}
\small

\begin{tabular}{lccccc}
\toprule
\textbf{Method} 
& \textbf{Sensor-Agnostic}
& \textbf{Object-Agnostic}
& \textbf{Missing Objects}
& \textbf{Completion Strategy}
& \textbf{Training} \\
\midrule

Global Rescaling \citep{Yang2024DepthV2} 
& \checkmark & \checkmark & $\times$ & Global linear rescaling & $\times$ \\

LWLR \citep{Xu2022Towards3S} 
& \checkmark & \checkmark & $\circ$ & Local weighted regression & $\times$ \\

PriorDA \citep{wang2025depthprior} 
& $\times$ & $\times$ & $\times$ & kNN coarse completion + MDE & \checkmark \\

Song et al. \citep{SONG2026155} 
& $\times$ & \checkmark & \checkmark & Random Forest regression & \checkmark \\

TomoSAR2Height \citep{chen2026tomosar2height} 
& $\times$ & $\times$ & $\circ$ & CNN-based reconstruction & \checkmark \\

\midrule
\textbf{Prior2DSM (Proposed)} 
& \checkmark & \checkmark & \checkmark & DINOv3 + MLP & $\times$ \\

\bottomrule

\end{tabular}
\end{table*}
\section{Related Work}

\subsection{Monocular height estimation}
Deep learning has significantly advanced MHE from RS imagery. Pioneering works, such as IM2HEIGHT, introduced fully convolutional networks to directly regress digital surface models (DSMs) from monocular images \citep{mou2018im2heightheightestimationsingle}. Subsequent CNN-based approaches further refined this direction: IMG2nDSM 
employed a residual-block architecture to estimate buildings and vegetation 
heights from single aerial RGB images, achieving strong performance, 
across urban and rural scenes \citep{rs13122417}. More recently, vision Transformers 
(ViTs) enabled improved spatial consistency by leveraging 
self-attention mechanisms to capture global contextual dependencies, 
motivating several ViT-based RS height estimation methods 
\citep{10294289}.

Complementing these domain-specific MHE architectures, foundation 
monocular depth estimation (MDE) models, used in computer vision fields, were recently adapted for  height estimation using RS data  \citep{cambrin2024depth}. 
Depth Anything V2 (DAV2), trained on large-scale mixtures of labeled 
and unlabeled visual data, achieved high performance on the DA-2K 
benchmark, across eight image categories, including aerial imagery 
(~9\% of the dataset) \citep{Yang2024DepthV2}. When fine-tuned for RS data, DAV2 demonstrates substantial improvements: up to 24\% 
RMSE reduction in urban height reconstruction \citep{hong2025depth2elevation} 
and state-of-the-art canopy height estimation with significantly reduced 
computational requirements \citep{cambrin2024depth}. Among available 
foundation models: Depth Anything V2 offers balanced performance and 
efficiency \citep{Yang2024DepthV2}; Depth Pro provides metric depth 
prediction without metadata \citep{bochkovskii2025depthprosharpmonocular}; 
while MoGe-2 emphasizes geometric consistency through multi-view 
constraints \citep{wang2025moge2accuratemonoculargeometry}. However, 
their performance for RS height estimation, remains ill-posed \citep{song2025enhancing}.

Despite these advances \citep{8128167}, MHE 
exhibit fundamental limitations under diverse imaging conditions. 
Near-orthographic viewing geometry eliminates perspective height cues, 
forcing reliance on shadow information for height inference 
\citep{SONG2026155, yu2026large}. This shadow-dependence induces 
substantial sensitivity: variations in illumination geometry can cause 
height errors exceeding 30\% \citep{ESFAHANI2025759}, resulting in  
dataset bias and poor generalization across domains 
\citep{zeng2024rsaresolvingscaleambiguities, hong2025depth2elevation}. 
These limitations have motivated physically constrained approaches that 
anchor MHE to sparse metric priors, enabling effective 
height completion, updating, and recovery 
\citep{guo2025multiviewreconstructionsfmguidedmonocular}.

\subsection{Height Completion}

Traditional height completion methods rely on spatial interpolation techniques such as kriging, inverse distance weighting, or geomorphologically constrained algorithms \citep{ajvazi2019comparative, 10604850}. While effective for filling gaps caused by sensor noise, data voids, or cloud occlusion, these methods assume spatial continuity of terrain surface \citep{reuter2007evaluation}. Consequently, they fail when entire surface features are 
absent from the prior, fundamentally limiting their application to DEM updating scenarios involving new construction, land-cover change, 
or vegetation regrowth \citep{kolzenburg2016rapid}.

More advanced completion methods use diffusion for void-filling. GAN-based approaches \citep{ZHAO2024114432, 8669876} improve upon classical interpolation by learning data-driven priors. Although effective in some settings, these approaches are computationally expensive and may introduce geometric distortions when sparse priors are weak, unevenly distributed, or inconsistent \citep{viola2025Marigolddczeroshotmonoculardepth}.

Therefore, recent studies have explored MDE as a complementary strategy for height completion, particularly when sparse metric priors are available \citep{11202549}. These methods leverage relative geometry from foundation models and align it to sparse height measurements through scale–shift transformations \citep{guo2025monocularoneshotmetricdepthalignment, SONG2026155}. This alignment is commonly performed using global affine fitting or local weighted linear regression (LWLR) \citep{Xu2022Towards3S}.

Recently, prior-based MDE models have gained increasing attention in the computer vision (CV) field as a means to address the ill-posed nature of monocular depth estimation \citep{hyoseok2025zeroshotdepthcompletiontesttime}. These approaches incorporate sparse or incomplete depth observations as conditioning inputs to pretrained MDE models, enabling the integration of measured geometric information into monocular inference \citep{yu2026large}. For example, PriorDA introduces a KNN-based coarse completion stage followed by a conditioned ViT-based MDE model, achieving state-of-the-art performance in completing missing regions of depth maps \citep{wang2025depthprior}. 

In parallel, test-time optimization methods have also demonstrated competitive performance without requiring dataset-specific training. Notably, Marigold-DC builds on a pretrained latent diffusion model (LDM) for depth estimation and incorporates sparse depth observations through test-time adaptation (TTA), enabling effective depth completion without additional training \citep{viola2025Marigolddczeroshotmonoculardepth}.

While prior-based MDE methods have been widely explored, research on monocular prior-based MHE remains more limited. In particular, Song et al. \citep{SONG2026155} introduced a sparse LiDAR-guided correction model based on Random Forest, trained on features extracted from Depth Anything V2 and RS3DAda to predict per-pixel rescaling factors. Their results demonstrate consistent improvements over global linear fitting. Additionally, TomoSAR2Height proposes a feature reconstruction strategy using a U-shaped deep convolutional network and achieves higher accuracy than bilinear interpolation for inpainting building structures \citep{chen2026tomosar2height}. 

Despite their strong performance, these approaches are inherently sensor-specific and dataset-dependent, and were evaluated only with individual monocular foundation model backbones rather than through systematic comparison. Consequently, their applicability remains limited, and they do not readily support sensor-agnostic or object-agnostic height completion. Table~\ref{tab:comparison} summarizes the key characteristics of existing approaches. These limitations motivate our training-free, feature-space completion approach that leverages self-supervised ViT representations for dense height propagation while systematically benchmarking multiple monocular foundation models for refinement.

\begin{figure*}[ht!]
    \centering
    \includegraphics[width=1\linewidth]{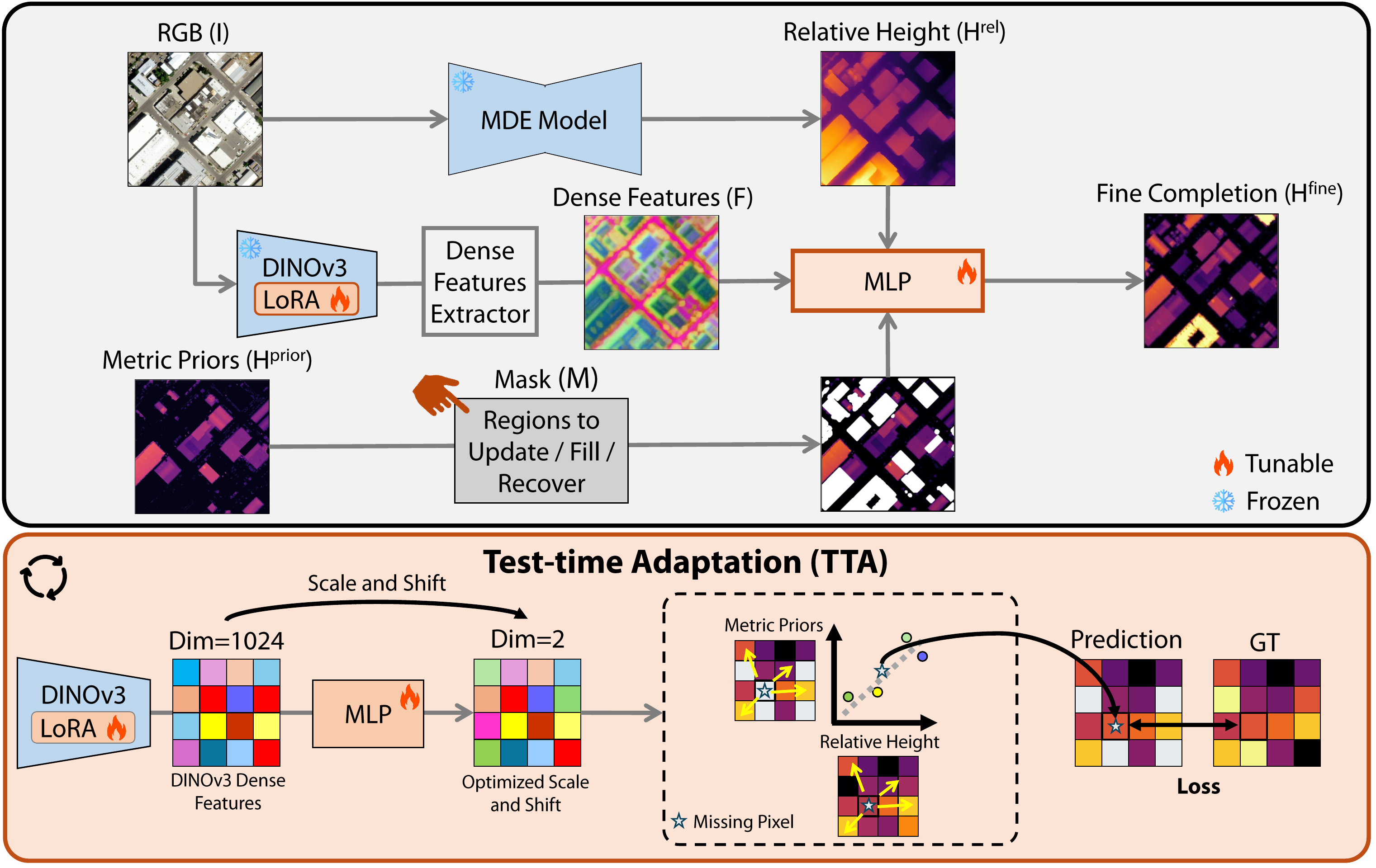}
    \caption{Overview of the proposed height completion framework. Given an RGB image, a monocular depth estimation (MDE) model produces a relative height map, while a self-supervised DINOv3 encoder extracts dense semantic features. Incomplete metric priors are used together with a change mask to identify regions that require completion. A lightweight MLP predicts spatially varying scale and shift parameters that transform relative height predictions into metric heights. During test-time adaptation (TTA), LoRA is applied to the DINOv3 attention layers and the MLP is optimized using the available metric priors, enabling local calibration of the relative height prediction. The final output is a refined metric DSM where missing regions are reconstructed while preserving structural consistency.}
    \label{framework}
\end{figure*}

\subsection{Deep Feature Similarity}

To address scale ambiguity in monocular estimation, recent methods incorporate semantic information, such as RGB features or text descriptions, which provide more consistent cues for metric alignment than geometric features alone \citep{11346819}. Moreover, pixels belonging to the same object, or material class, tend to share similar depth distortions even when spatially separated \citep{Jiao_2018_ECCV}. RGB features were proven effective for propagating depth information from sparse measurements to dense predictions 
\citep{yu2026large, lin2025promptingdepth4kresolution}. Furthermore, textual object descriptions improve relative-to-metric depth alignment, achieving accuracy comparable to oracle affine fitting directly to ground truth \citep{zeng2024rsaresolvingscaleambiguities}. However, despite their effectiveness, these methods still rely on training dedicated RGB or text encoders \citep{cui2025tr2mtransferringmonocularrelative}.

Recently, self-supervised Vision Transformers (ViTs) demonstrated high potential for zero-shot semantic correspondence, offering robust and transferable representations without task-specific supervision \citep{oquab2024dinov2learningrobustvisual}. Among these models, DINO scales self-supervised learning to produce universal ViT backbones with strong transferability across downstream tasks \citep{siméoni2025dinov3}. DINO features encode rich, well-localized semantic information at high spatial granularity, such as object parts, and share semantic representations across related, yet distinct, object categories \citep{amir2022deepvitfeaturesdense}.

DINO features demonstrate semantic consistency across object 
categories in natural images \citep{siméoni2025dinov3, 
amir2022deepvitfeaturesdense}. 

We \textit{hypothesize} here, that under orthographic viewing conditions, common in RS imagery, similar structural elements—buildings, vegetation, and terrain of similar types—produce consistent self-attention patterns regardless of spatial location, enabling semantic-guided height transfer. This suggests that scale and shift parameters can be inferred, at the semantic-object level, without relying on explicit priors or retraining, enabling dense per-pixel estimation.

\section{Materials and Methodology}

\subsection{Problem Setting}

We consider a scenario in which incomplete or outdated height information is available together with up-to-date monocular imagery. Let $\mathbf{H}^{\mathrm{prior}} \in \mathbb{R}^{H \times W}$ denote a prior height map that may contain correct metric values in unchanged regions and erroneous values in locations where structures were missed, modified, distorted, or removed.

Let $\mathbf{I} \in \mathbb{R}^{H \times W \times 3}$ denote the corresponding RGB image capturing the complete and updated surface appearance. In addition, we assume the availability of a binary change mask 
$\mathbf{M} \in \{0,1\}^{H \times W}$, where $\mathbf{M}(x,y)=1$ indicates pixels affected by temporal changes and $\mathbf{M}(x,y)=0$ denotes stable regions.

The objective is to reconstruct a complete and updated height map 
$\mathbf{H}^{\mathrm{fine}} \in \mathbb{R}^{H \times W}$ that accurately represents the current surface geometry by integrating prior height observations with the updated monocular imagery.

To achieve this, the proposed framework rescales the monocular depth estimation output $\mathbf{H}_{\mathrm{rel}}$ at the pixel level. Missing or unreliable regions in $\mathbf{H}^{\mathrm{prior}}$ are recovered through test-time optimization of spatially varying scale and shift parameters. This optimization is guided by semantic feature representations $\mathbf{F}$ and monocular relative height predictions $\mathbf{H}_{\mathrm{rel}}$, enabling the fusion of relative height estimates with the prior map. The resulting reconstruction produces a detailed and metrically consistent height map $\mathbf{H}^{\mathrm{fine}}$ (Fig.~\ref{framework}).

\subsection{Depth Foundation Models}

Depth foundation models \citep{Yang2024DepthV2, birkl2023midas} follow an encoder--decoder architecture. They are developed upon multi-stage feature extraction using a ViT-L encoder with a pretrained DINOv2 backbone, coupled with a convolution-based decoder, following the Dense Prediction Transformer (DPT) framework \citep{ranftl2021vision}.

While Depth Anything~V2 (DAV2) demonstrates robustness for general scene understanding \citep{Yang2024DepthV2}, it tends to underperform compared with recent architectures, such as Depth Pro and MoGe-2, in capturing high-frequency details and preserving geometric sharpness. Depth Pro employs a multi-scale ViT backbone that produces sharper object boundaries, and better retains fine-grained structures \citep{bochkovskii2025depthprosharpmonocular}. Similarly, MoGe-2 emphasizes edge consistency and explicit 3D geometric alignment, yielding a more stable structural representation \citep{wang2025moge2accuratemonoculargeometry}.

\subsection{Dense Feature Extraction}

Standard Vision Transformer (ViT) architectures are limited by a rigid $P \times P$ patch grid, which often results in "blocky" semantic maps. To resolve fine architectural details, we implement a strided overlap accumulation strategy. By shifting the input image $\mathbf{I}$ by a sub-patch stride $s$ (where $s < P$), we extracted multiple overlapping semantic views for every pixel. 

\begin{figure}[ht!]
    \centering
    \includegraphics[width=1\linewidth]{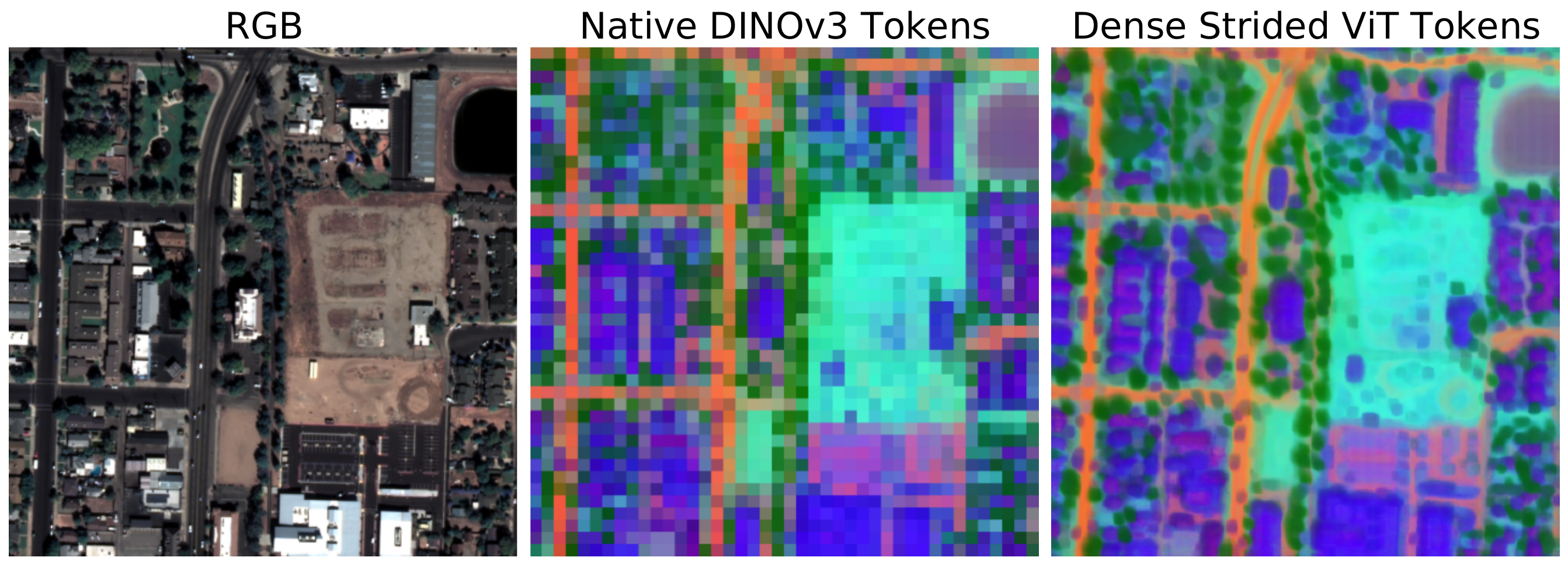}
    \caption{Comparison of semantic feature representations extracted from DINOv3. Left: input RGB aerial image. Middle: native DINOv3 patch tokens obtained using the standard patch grid. Right: dense strided ViT tokens obtained using overlapping patch extraction, significantly increasing spatial sampling density and yielding smoother, higher-resolution semantic feature maps that better preserve fine spatial structures.}
    \label{semantic_resolution_comparison}
\end{figure}

The final dense feature at any coordinate is defined as the average of all patch tokens that cover that specific location. This ensembling approach effectively yields a 16-fold increase in semantic sampling density (for $s=4, P=16$), see the comparison in Fig. \ref{semantic_resolution_comparison}. 

\subsection{Tunable Components}

\textit{Low-rank adaptation}:
To enable efficient adaptation (LoRA) of DINOv3 without updating its full set of parameters, we employ LoRA to the attention projection layers. In our implementation, LoRA is injected into both the combined \textit{qkv} projection and the attention output projection. Given an input token sequence $X \in \mathbb{R}^{L \times d_c}$, the adapted projection is expressed as
\begin{equation}
W'_m = W_m + \Delta W_m ,
\end{equation}
where $W_m$ denotes the frozen pretrained weight matrix and $\Delta W_m$ represents a learnable low-rank update. Following \citep{hu2021loralowrankadaptationlarge}, the update is parameterized as
\begin{equation}
\Delta W_m = \frac{\alpha}{r} B_m A_m ,
\end{equation}
where $A_m \in \mathbb{R}^{r \times d_{\mathrm{in}}}$ and $B_m \in \mathbb{R}^{d_{\mathrm{out}} \times r}$ are trainable low-rank matrices and $r$ denotes the adaptation rank. In our experiments we set $r=8$ and $\alpha=16$. During optimization, the original pretrained weights remain frozen and only the LoRA parameters are updated. This parameter-efficient fine-tuning strategy enables the model to adapt attention projections to the target scene while preserving the general visual representations learned during large-scale self-supervised pretraining.

\textit{MLP scale and shift}: To convert relative depth estimates into metric heights, we introduce a lightweight three-layer perceptron (MLP) that predicts a spatially varying affine transformation for each strided transformer token. 
Given the token embeddings extracted from the vision transformer, the MLP outputs two parameters per token: a multiplicative scale $s$ and an additive bias $b$. 
These parameters define an affine calibration of the relative depth prediction $r$ so that the metric height is estimated as $\hat{z} = s \cdot r + b$. 
The MLP consists of two hidden layers with nonlinear activations and layer normalization, enabling the network to learn a mapping from semantic token features to local height calibration parameters. 
During test-time optimization, MLP parameters are adjusted using incomplete anchor heights, allowing the model to locally correct systematic scale and bias errors in the relative depth prediction while maintaining spatial consistency across the scene, finally producing a high-resolution metric height map $\mathbf{H}_{\mathrm{fine}}$.

\subsection{Dataset}

We assembled a multi-source dataset from publicly available repositories to  evaluate Prior2DSM. The dataset consists of two subsets compiled from four primary data sources:
(1) high-resolution airborne RGB imagery of Denver acquired through the National Agriculture Imagery Program (NAIP) \citep{usgs_naip}, with a spatial resolution of $30\,\text{cm} \times 30\,\text{cm}$ per pixel;
(2) high-resolution WorldView-3 (WV3) satellite RGB imagery of Fresno city, with a spatial resolution of $60\,\text{cm} \times 60\,\text{cm}$ per pixel;
(3) high-resolution LiDAR-derived digital surface models (DSM) obtained from the OpenTopography platform \citep{opentopography}; and
(4) land use and land cover (LULC) data.

The Denver NAIP coverage spans approximately $22\,\text{km}^2$, while the Fresno WV3 subset covers approximately $12\,\text{km}^2$. Both datasets were divided into non-overlapping $672 \times 672$ patches. The NAIP dataset includes the city of Denver which encompasses residential and industrial regions, including taller structures and complex urban morphology, resulting in higher height variability. Specifically, exhibits a mean ground-truth height of 4.22\,m with a standard deviation of 7.01\,m. In contrast, WV3 data capture Fresno residential regions which predominantly contain low-rise houses with more homogeneous and repetitive structures, yielding a lower mean height of 2.21\,m and a standard deviation of 2.80\,m.

\begin{figure*}[ht!]
    \centering
    \includegraphics[width=1\linewidth]{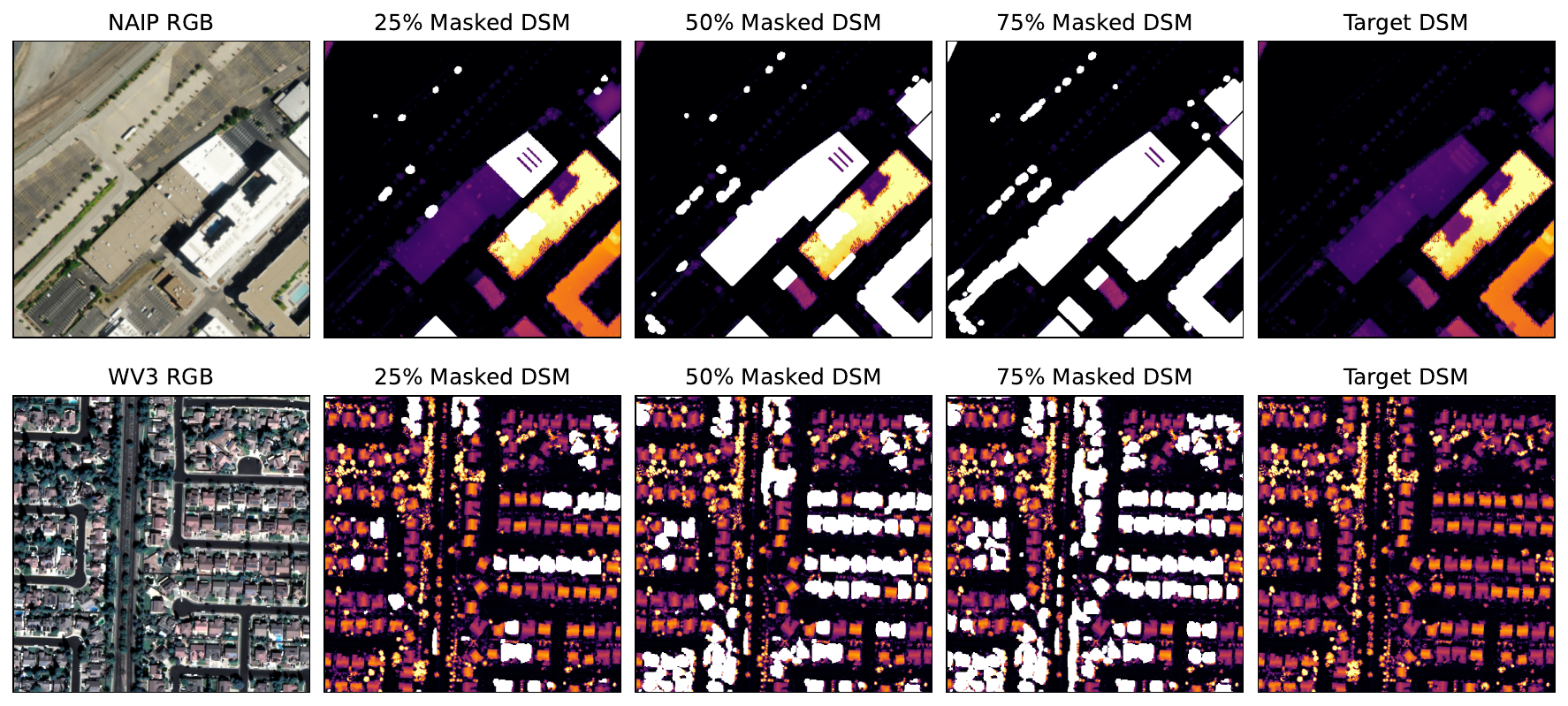}
    \caption{The experimental dataset consists of two domains: (1) the Denver NAIP dataset and (2) the Worldview-3 Fresno dataset. Both datasets include high-resolution RGB imagery, ground-truth normalized DSM (nDSM), and three levels of prior incompleteness (25\%, 50\%, and 75\%) for evaluation.
 }
    \label{datasets}
\end{figure*}

\subsection{Experiment Design}

To design the main experiment, we evaluate Prior2DSM on the task of metric height completion under progressively increasing data removal. Specifically, for each of the two height datasets, we remove $25\%$, $50\%$, and $75\%$ of the metric height observations by converting them to no data cells, see Fig. \ref{datasets}. Missing regions are generated using land use and land cover (LULC) information by randomly selecting surface objects (e.g., buildings or vegetation) and masking them together with a surrounding buffer of $10$~m. This buffer ensures that both the selected objects, and adjacent ground regions, are fully removed, preventing partial object leakage and enforcing a true completion scenario. The resulting random object-missing masks are stored and reused consistently across all compared methods.

We compare the proposed coarse alignment strategy against several representative baselines to assess robustness and accuracy under sparse metric supervision. Specifically, we evaluate:  
(i) \emph{bilinear interpolation}, the filling strategy employed in PromptDA;  
(ii) \emph{global relative rescaling}, which estimates a single scale--shift pair using all available metric pixels and applies it uniformly across the image;  
(iii) \emph{LWLR rescaling}, which performs locally weighted linear regression for each missing pixel based on nearby metric observations; and  
(iv) \emph{Spatial $k$NN alignment}, where a local scale--shift pair is estimated using the $K$ nearest sparse metric points in the spatial domain \citep{wang2025depthprior}.

We further compare our prior based framework with globally rescaled non-prior monocular height estimation RS3Dada model \citep{song2024synrs3dsyntheticdatasetglobal}. This comparison enables an assessment of framework compared with a pure robust monocular model.

Beyond masking experiments, we evaluate the framework on a real-world temporal inconsistency scenario using an outdated DSM from Denver acquired in 2008. In this setting, outdated regions are identified by subtracting two ground-truth DSMs acquired at different times and masking regions corresponding to newly constructed or modified structures. The resulting missing regions are completed using the same protocol as in the main experiment, enabling an end-to-end evaluation under realistic temporal discrepancies.

In addition to robustness testing, we coupled Prior2DSM with the Text2Earth RS inpainting model and provide visual examples of DSM generation. This comparison demonstrates the ability of Prior2DSM to successfully process synthetic signals and highlights its capability for practical, detailed DSM simulation across applications.

Finally, we conduct three ablation experiments to analyze the contribution of the key components of Prior2DSM. First, we predict from dense DINOv3 features direct height with MLP (without rescaling MDE). Second, we replace the Depth Anything V2 backbone used for coarse relative alignment with alternative relative depth foundation models, including \emph{Depth Pro} and \emph{MoGe-2}, to assess the robustness of the framework with respect to backbone selection. Finally, we investigate the impact of the LoRA of DINOv3 by optimize only the MLP scale and shift predictor during the inference.

\subsection{Evaluation metrics}
Two commonly used height error evaluation metrics, the mean absolute error (MAE) and the root mean square error (RMSE), are used, as shown in Eqs.~\ref{eq:mae}–\ref{eq:ssim}

\begin{equation}
\text{MAE} = \frac{1}{N} \sum_{i=1}^{N} \left| y_i - \hat{y}_i \right|,
\label{eq:mae}
\end{equation}

\begin{equation}
\text{RMSE} = \sqrt{ \frac{1}{N} \sum_{i=1}^{N} (y_i - \hat{y}_i)^2 },
\label{eq:rmse}
\end{equation}

To assess structural fidelity, we further report the Structural Similarity Index Measure (SSIM), which evaluates perceptual similarity by jointly considering local luminance, contrast, and structural consistency between the reconstructed and ground-truth height maps:

\begin{equation}
\text{SSIM}(y, \hat{y}) =
\frac{(2\mu_y \mu_{\hat{y}} + C_1)(2\sigma_{y\hat{y}} + C_2)}
{(\mu_y^2 + \mu_{\hat{y}}^2 + C_1)(\sigma_y^2 + \sigma_{\hat{y}}^2 + C_2)},
\label{eq:ssim}
\end{equation}
where $\mu_y$ and $\mu_{\hat{y}}$ denote the mean values of the ground-truth and predicted height maps, respectively; $\sigma_y^2$ and $\sigma_{\hat{y}}^2$ denote their variances; $\sigma_{y\hat{y}}$ denotes the covariance between $y$ and $\hat{y}$; and $C_1$ and $C_2$ are small constants introduced to stabilize the division.

\begin{table*}[t]
\centering
\caption{Mean metrics of NAIP dataset against GT for 25\%, 50\%, and 75\% completion.}
\setlength{\tabcolsep}{1.5 pt}

\begin{tabular}{lccccccccc}
\toprule
\textbf{\multirow{4}{*}{}} &
\multicolumn{3}{c}{25\% Completion} &
\multicolumn{3}{c}{50\% Completion} &
\multicolumn{3}{c}{75\% Completion} \\
\cmidrule(lr){2-4} \cmidrule(lr){5-7} \cmidrule(lr){8-10}
& MAE (m) $\downarrow$ & RMSE (m) $\downarrow$ & SSIM $\uparrow$
& MAE (m) $\downarrow$ & RMSE (m) $\downarrow$ & SSIM $\uparrow$
& MAE (m) $\downarrow$ & RMSE (m) $\downarrow$ & SSIM $\uparrow$ \\
\midrule

Global Rescaling  (Baseline)      
& 5.08 & 9.01 & 0.82
& 4.81 & 8.45 & 0.82
& 4.74 & 8.57 & 0.82 \\

Bilinear Interpolation          
& 4.30 & 8.42 & 0.80
& 4.72 & 9.16 & 0.78
& 5.06 & 9.75 & 0.75 \\

LWLR Rescaling             
& 4.93 & 8.97 & 0.80
& 4.70 & 8.63 & 0.81
& 4.57 & 8.42 & 0.82 \\

Spatial kNN  Rescaling
& 3.88 & 7.75 & 0.84
& 4.25 & 8.49 & 0.82
& 4.45 & 8.64 & 0.80 \\

\midrule
RS3DAda
& 4.07 & 7.80 & 0.87
& 3.85 & 7.20 & 0.88
& 3.80 & 7.16 & 0.88  \\

Marigold-DC             
& 3.66 & 6.59 & 0.86
& 4.09 & 7.61 & 0.84
& 4.49 & 8.23 & 0.80 \\

PriorDA + DAv2            
& 3.39 & 5.95 & 0.88
& 3.74 & 6.79 & 0.86
& 4.17 & 7.57 & 0.83 \\

\midrule

Prior2DSM + DAv2 
& 2.75 & 5.49 & 0.92
& 2.94 & 6.03 & 0.91
& 3.28 & 6.70 & 0.89 \\

Prior2DSM + Depth Pro              
& 2.73 & 5.36 & 0.92
& 2.91 & 5.87 & 0.91
& 3.25 & 6.67 & 0.89 \\

Prior2DSM + MoGe-2              
& \textbf{2.71} & \textbf{5.31} & \textbf{0.92}
& \textbf{2.89} & \textbf{5.85} & \textbf{0.91}
& \textbf{3.24} & \textbf{6.59} & \textbf{0.89} \\

\bottomrule
\end{tabular}
\label{tab:NAIP_MAIN}
\end{table*}

\begin{table*}[t]
\centering
\caption{Mean metrics of WV3 dataset against GT for 25\%, 50\%, and 75\% completion.}
\setlength{\tabcolsep}{1.5 pt}

\begin{tabular}{lccccccccc}
\toprule
\textbf{\multirow{4}{*}{}} &
\multicolumn{3}{c}{25\% Completion} &
\multicolumn{3}{c}{50\% Completion} &
\multicolumn{3}{c}{75\% Completion} \\
\cmidrule(lr){2-4} \cmidrule(lr){5-7} \cmidrule(lr){8-10}
& MAE (m) $\downarrow$ & RMSE (m) $\downarrow$ & SSIM $\uparrow$
& MAE (m) $\downarrow$ & RMSE (m) $\downarrow$ & SSIM $\uparrow$
& MAE (m) $\downarrow$ & RMSE (m) $\downarrow$ & SSIM $\uparrow$ \\
\midrule

Global Rescaling (Baseline)   
& 2.06 & 2.88 & 0.91 
& 2.11 & 2.91 & 0.91 
& 2.20 & 3.02 & 0.90  \\

Bilinear Interpolation  
& 2.08 & 2.91 & 0.90 
& 2.14 & 2.99 & 0.89 
& 2.18 & 3.03 & 0.89  \\

LWLR Rescaling
& 2.81 & 3.91 & 0.84 
& 2.98 & 4.09 & 0.82 
& 3.05 & 4.23 & 0.82  \\

Spatial kNN Rescaling
& 2.07 & 3.29 & 0.91 
& 2.09 & 3.28 & 0.91 
& 2.15 & 3.39 & 0.91  \\

\midrule

RS3DAda
& 1.95 & 2.77 & 0.92
& 1.98 & 2.76 & 0.92
& 2.03 & 2.81 & 0.91  \\

Marigold-DC
& 1.88 & 2.82 & 0.92
& 2.04 & 3.01 & 0.91
& 2.35 & 3.57 & 0.89 \\ 

PriorDA + DAv2
& 2.01 & 2.72  & 0.92
& 2.08 & 2.79  & 0.91
& 2.18 & 2.92  & 0.90  \\

\midrule

Prior2DSM + DAv2
& \textbf{1.55} & \textbf{2.36} & \textbf{0.94}
& 1.63 & 2.46 & 0.94 
& \textbf{1.77} & \textbf{2.60} & \textbf{0.93}  \\

Prior2DSM + Depth Pro
& 1.57 & 2.38 & 0.94
& \textbf{1.63} & \textbf{2.45} & \textbf{0.94}
& 1.78 & 2.61 & 0.93 \\

Prior2DSM + MoGe-2
& 1.56 & 2.38 & 0.94
& 1.63 & 2.46 & 0.94
& \textbf{1.77} & \textbf{2.60} & \textbf{0.93} \\

\bottomrule
\end{tabular}
\label{tab:wv3_main}
\end{table*}

\begin{table}[t]
\centering
\caption{Ablation study of Prior2DSM (mean results across both datasets for 50\% completion).}
\setlength{\tabcolsep}{0.8pt}

\begin{tabular}{lccc}
\toprule
\textbf{Method} 
& MAE (m) $\downarrow$ 
& RMSE (m) $\downarrow$ 
& SSIM $\uparrow$ \\
\midrule

w/o MDE (MLP $\rightarrow$ Height)
& 2.71
& 5.52
& 0.91 \\

w/o LoRA (Frozen DINOv3)
& 2.66
& 5.32
& 0.91 \\

Prior2DSM
& \textbf{2.40}
& \textbf{4.97}
& \textbf{0.92} \\

\bottomrule
\end{tabular}

\label{tab:ablation_prior2dsm}
\end{table}

\begin{figure*}[ht!]
    \centering
    \includegraphics[width=1\linewidth]{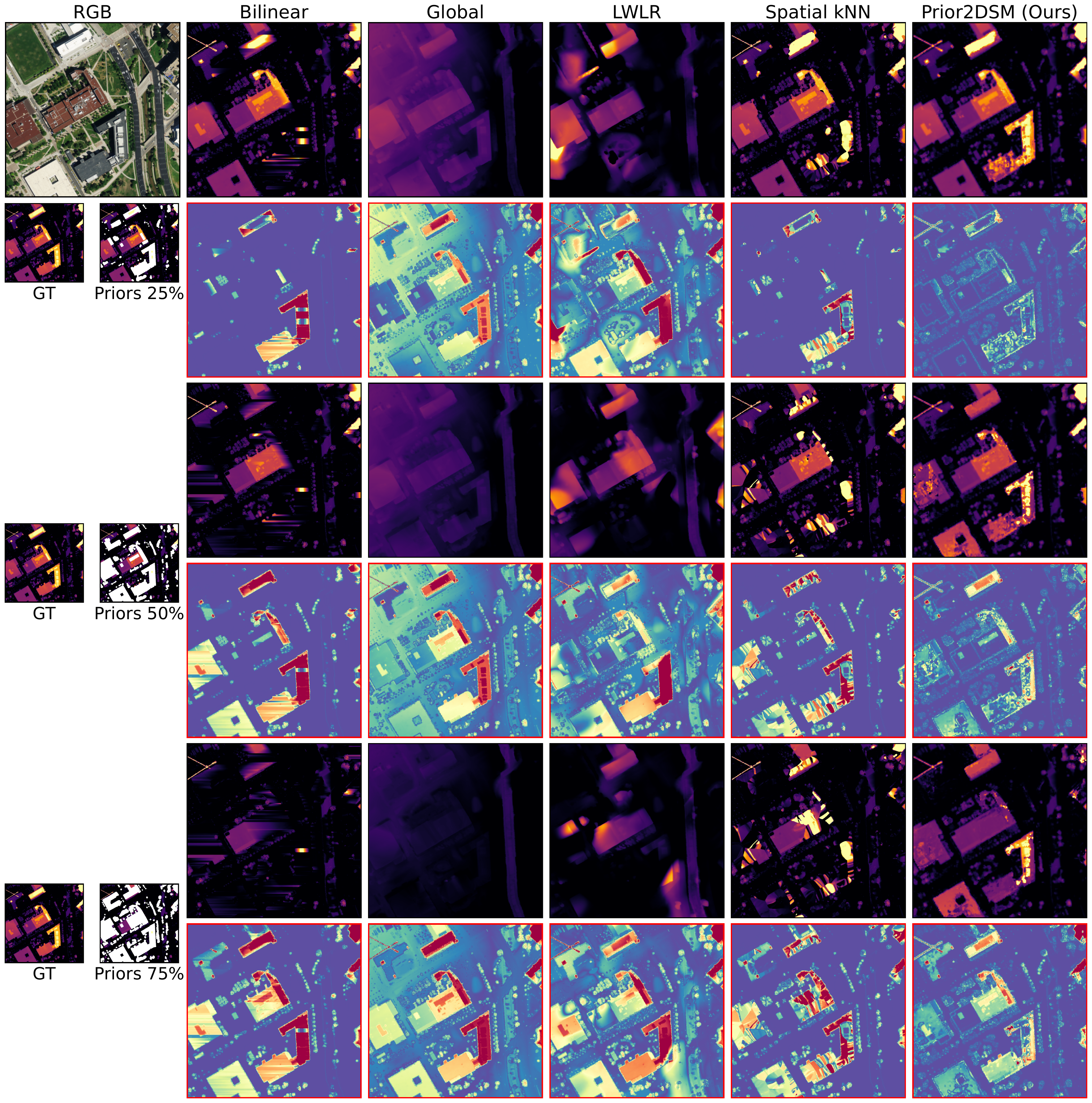}
    \caption{Comparison of DSM completion methods. From left to right: RGB image, Bilinear interpolation, Global affine fitting, Locally Weighted Linear Regression (LWLR), Spatial kNN propagation, DINOv3 feature-space kNN, and the proposed Prior2DSM. Each block corresponds to different levels of missing prior data (25\%, 50\%, and 75\%). For each case, the ground truth (GT) DSM and the degraded prior input are shown on the left, followed by the reconstruction results and the corresponding absolute error maps. While simple interpolation and global fitting produce oversmoothed surfaces and spatial artifacts, semantic feature-based approaches better preserve structural details. The proposed framework consistently achieves more accurate reconstructions with lower errors, particularly under severe missing data conditions. Height: black = low, yellow = high. Error: blue = low, red = high.}
    \label{main_visual}
\end{figure*}

\section{Results}

Tables~\ref{tab:NAIP_MAIN} and \ref{tab:wv3_main} summarize the quantitative results on NAIP and WV3 datasets for metric completion, under three levels of prior incompleteness, evaluated in terms of MAE, RMSE, and SSIM on completed pixels. As expected, error consistently increases as the level of incompleteness rises from 25\% to 75\% for all methods. We compared Prior2DSM to non-parametric methods, including bilinear interpolation, local and global rescaling, as well as spatial kNN.

The reported metrics correspond only to completed pixels (i.e., previously missing regions). For the NAIP dataset, bilinear interpolation, LWLR, and global rescaling yielded the highest errors. At 25\% incompleteness, global rescaling underperforms with nearly 9\,m RMSE compared to 8.42\,m for bilinear interpolation. At higher levels of incompleteness, global rescaling slightly outperforms bilinear interpolation. Visual inspection shows that bilinear interpolation fails to reconstruct fully missing objects, while LWLR introduces severe distortions in regions with missing data. In contrast, global rescaling produces geometrically cleaner results; however, in complex and highly structured urban areas such as Denver, linear fitting is insufficient to accurately rescale relative depth, leading to significant outliers and consistently high RMSE values (8--9\,m) across all levels of incompleteness.

Pixel-alignment-based methods perform better overall. We compare spatial kNN and DAv2-based rescaling from the PriorDA framework~\citep{wang2025depthprior}, which rescales missing pixels using the closest $K$ known neighbors.

The proposed framework yields the best performance among all evaluated rescaling and alignment strategies across all levels of incompleteness. Although the error increases as the missing ratio grows from 25\% to 75\%, the relative performance gain compared with other methods is most significant at 25\% incompleteness.

Specifically, the best performance on completed pixels in the NAIP dataset is achieved at 25\% incompleteness, with an MAE of 2.71\,m and an RMSE of 5.31\,m. At 50\% incompleteness, the RMSE is 5.85\,m, and at 75\% missing data, the error increases to 6.59\,m. In contrast, the WV3 dataset exhibits more stable behavior as incompleteness increases.

Overall, the error is higher in the NAIP dataset due to the characteristics of the capture area, particularly Denver, which contains more commercial and high-rise structures. Consequently, the performance gain in Denver is more pronounced. For example, at 25\% incompleteness, Prior2DSM achieves nearly a 46\% improvement in RMSE compared to global rescaling (9.91\,m vs.\ 5.31\,m), whereas in WV3 (Fresno) the improvement is approximately 18\% (2.88\,m vs.\ 2.36\,m). Visual comparisons between rescaling and alignment strategies are shown in Fig.~\ref{main_visual}.

Tables~\ref{tab:NAIP_MAIN} and \ref{tab:wv3_main} also provide comparison of Prior2DSM with state-of-the-art depth completion approaches, specifically PriorDA~\citep{wang2025depthprior} and Marigold-DC~\citep{viola2025Marigolddczeroshotmonoculardepth}. Our method consistently outperforms both in terms of MAE, RMSE, and SSIM.

For the NAIP dataset at 25\% incompleteness, PriorDA achieves an RMSE of 5.95\,m compared with 5.31\,m obtained by Prior2DSM, both using DAv2 as the relative depth source. Under more sparse prior conditions, PriorDA and Marigold-DC exhibit significantly higher reconstruction errors, whereas the proposed method is less sensitive to missing data. For example, at 75\% incompleteness, PriorDA reaches an RMSE of 7.57\,m, whereas Prior2DSM achieves 6.59\,m.

\begin{figure}[ht!]
    \centering
    \includegraphics[width=1\linewidth]{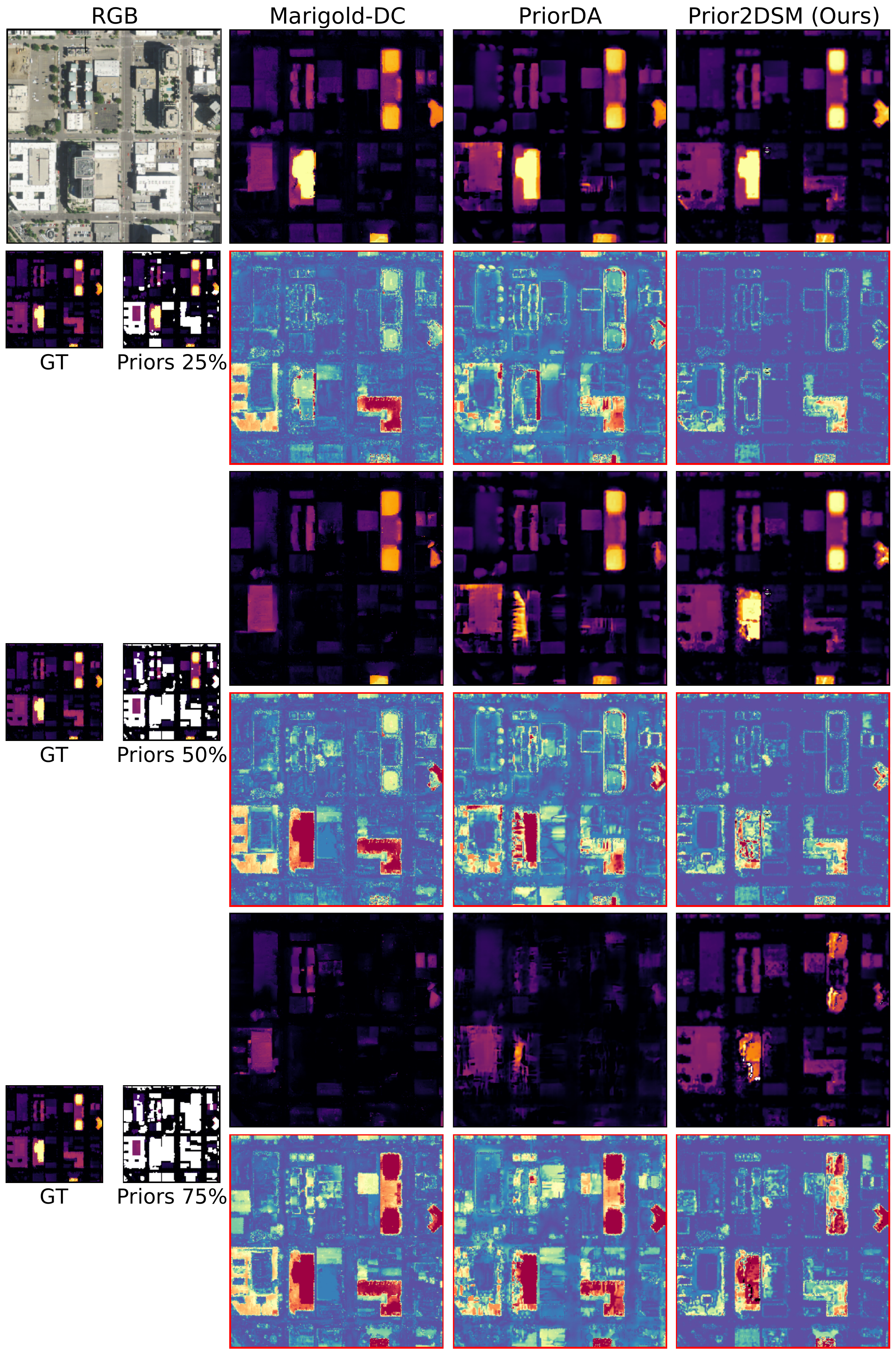}
    \caption{Prior-based DSM completion results: From left to right: RGB image, Marigold-DC, PriorDA, and the proposed framework. Each row corresponds to increasing levels of missing data (25\%, 50\%, and 75\%). For each case, the ground truth (GT) DSM and the degraded prior input are shown on the left. The second row in each block shows the corresponding absolute error maps. The framework Prior2DSM produces sharper building structures and lower reconstruction errors compared to existing prior-based approaches, particularly under severe missing data conditions. Height: black = low, yellow = high. Error: blue = low, red = high.}
    \label{visual_depth_sota}
\end{figure}

When compared with non-prior-based monocular depth models such as RS3DAda~\citep{song2024synrs3dsyntheticdatasetglobal}, RS3DAda yields an RMSE of approximately 7.04\,m on changed pixels in the NAIP dataset and 2.86\,m on the WV3 dataset. Prior2DSM outperforms this method at all levels of incompleteness, including under very sparse priors (75\% missing data). Visual comparisons with state-of-the-art methods are presented in Fig.~\ref{visual_depth_sota}.

\subsection{Ablation}

\textit{LoRA and MDE}: Table~\ref{tab:ablation_prior2dsm} summarizes the quantitative results of the proposed framework compared with two ablation variants. Direct MLP height prediction yields less accurate results than the MLP scale-and-shift prediction used in Prior2DSM (RMSE: 5.52 vs.\ 4.97). In addition, using a frozen DINOv3 backbone without LoRA also results in higher error (RMSE: 5.32).

\textit{Backbone Sensitivity of Relative MDE}: Tables~\ref{tab:NAIP_MAIN} and \ref{tab:wv3_main} summarize the quantitative results of Prior2DSM using three relative depth foundation models, which provide the relative geometry maps that define the relationship between priors and missing pixels.

On the more complex NAIP dataset, MoGe-2 consistently achieves better performance, while Depth Pro maintains lower RMSE values compared to DAv2.

\begin{figure}[ht!]
    \centering
    \includegraphics[width=1\linewidth]{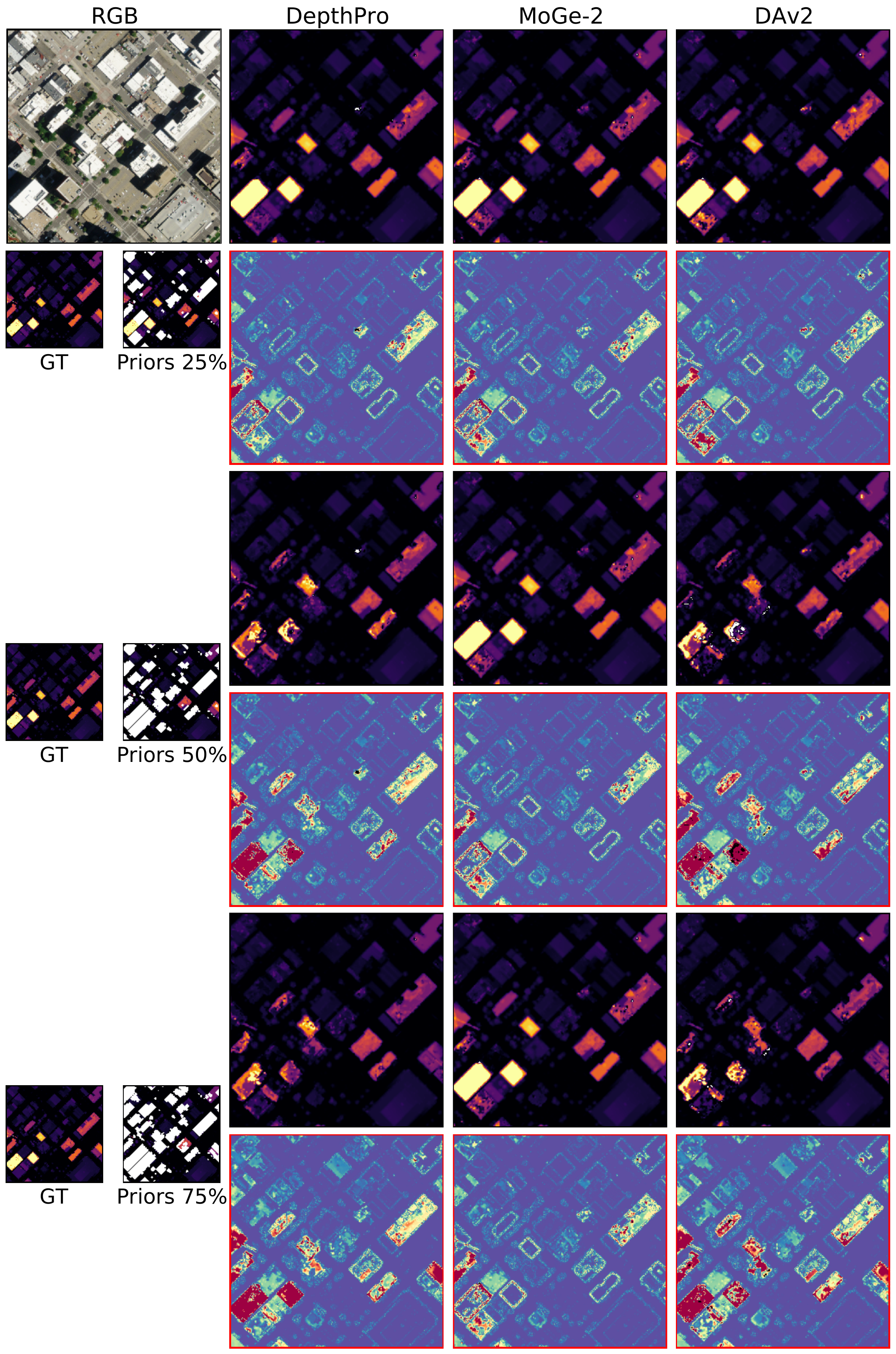}
    \caption{Visual comparison of relative height predictions from three models, Depth Anything V2 (DAV2), Depth Pro, and MoGe-2. DAV2 provides a smoother representation that better preserves flat regions such as ground and roads, while Depth Pro and MoGe-2 produce sharper and more detailed predictions with enhanced structural boundaries. Height: black = low, yellow = high. Error: blue = low, red = high.}
    \label{visual_depth_source}
\end{figure}

On the WV3 dataset, the differences between models are less pronounced. At 25\% completion, DAv2 achieves the lowest RMSE (2.36), followed closely by MoGe-2 and Depth Pro. At higher completion levels, all models perform almost equally. Examples of the different relative depth sources can be visually inspected in Fig.~\ref{visual_depth_source}.

\subsection{Application}
Table~\ref{tab:2008_results} summarizes the quantitative results for DSM updating using the 2008 elevation data. The original outdated DSM exhibits an RMSE of 5.35~m. After applying the proposed framework, the RMSE is reduced to 3.83~m, demonstrating a substantial improvement in metric accuracy.

\begin{figure}[ht!]
    \centering
    \includegraphics[width=1\linewidth]{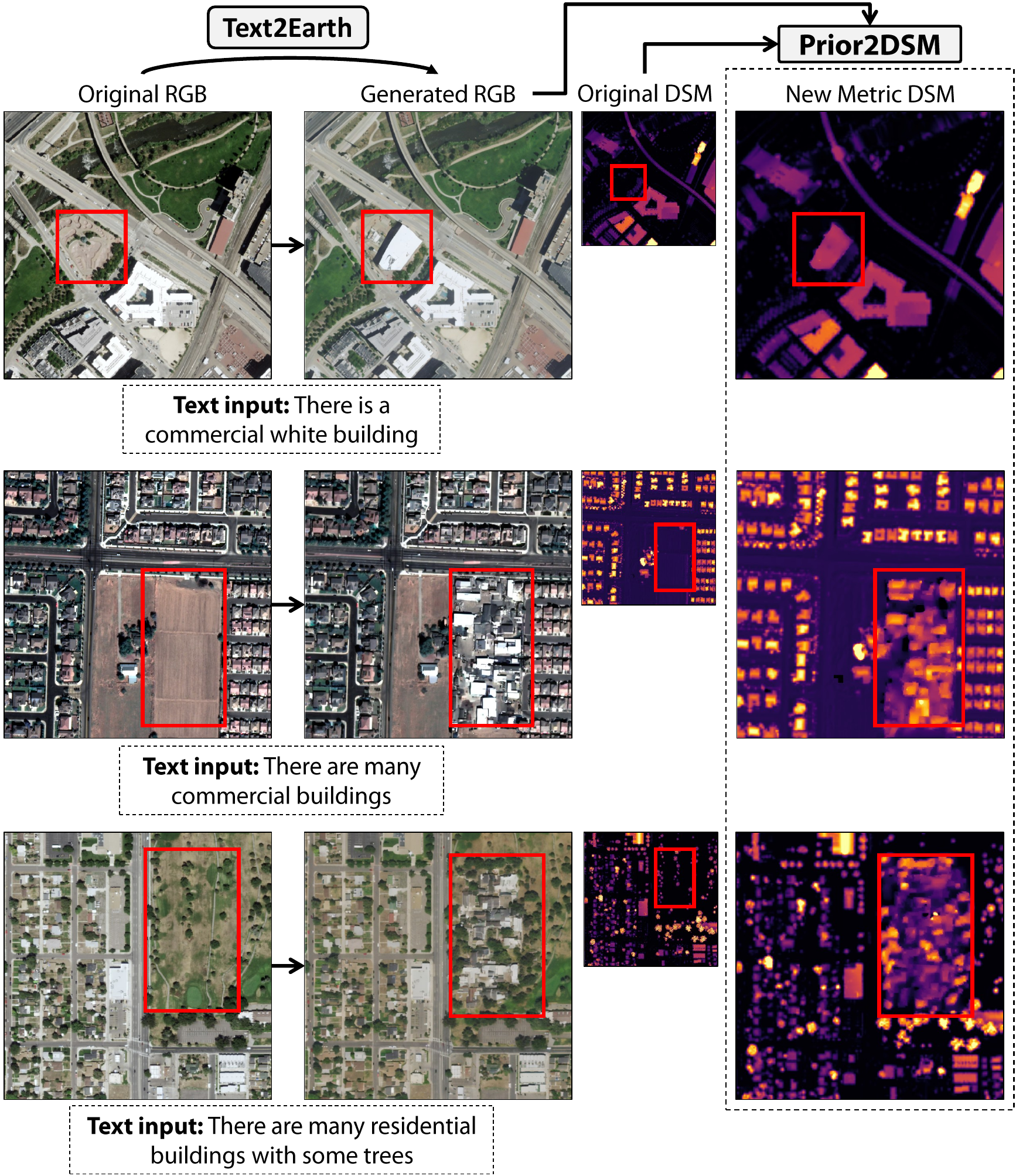}
    \caption{Coupled RGB–DSM editing using generative remote sensing models: Text2Earth generates modified RGB imagery from textual prompts describing new urban structures. The proposed framework then predicts the corresponding metric DSM for the edited scene. Red boxes indicate the regions where new buildings are introduced and reconstructed in the generated height map.}
    \label{text2earth}
\end{figure}

In addition to DSM updating, we further demonstrate metric DSM editing through a generative remote sensing image editing model. In Figure \ref{text2earth}, the newly generated metric DSM follows the edited RGB image content. Specifically, we showcase the capability of Prior2DSM as a modular component within an RGB--DSM co-editing Text2Earth framework \citep{liu2025text2earthunlockingtextdrivenremote}. Visually, the results indicate a seamless and metrically consistent updated DSM that realistically represents the newly generated features, such as commercial and residential buildings and trees \ref{text2earth}.

\begin{table}[t]
\centering
\caption{DSM updating results for the NAIP dataset.}
\setlength{\tabcolsep}{3pt}
\begin{tabular}{lccc}
\toprule
\textbf{Method} &
MAE (m) $\downarrow$ &
RMSE (m) $\downarrow$ &
SSIM $\uparrow$ \\
\midrule
None (DSM 2008)     & 1.89  & 5.35  & 0.92   \\
Prior2DSM       & 1.52 & 3.83 & 0.94 \\
\bottomrule
\end{tabular}
\label{tab:2008_results}
\end{table}

\section{Discussion} 

\subsection{Test-time adaptation  (TTA)}

Recent height completion models address the problem by designing and training task-specific networks that explicitly use a sparse depth map as an additional input channel \citep{tang2024bilateralpropagationnetworkdepth}. Pretrained models, however, suffer two key limitations. First, it is impractical that encompass the full diversity of geometric configurations, sparsity patterns, and sensor types \citep{hyoseok2025zeroshotdepthcompletiontesttime}. As a result, the learned encoder representations may overfit to the training distribution and struggle to generalize to unseen input conditions \citep{huang2025enhancinggeneralizationdepthestimation}. Second, adapting these models to new scenarios through retraining or fine-tuning is computationally demanding and may compromise the pretrained representations that enable their general applicability \citep{LUO2025126}.

TTA optimization allows the model to retain the broad geometric prior encoded in the base network while adapting to sparse evidence available during inference. TTA depth completion has recently demonstrated strong performance \citep{park2024testtimeadaptationdepthcompletion}, often outperforming pretrained depth completion models \citep{viola2025Marigolddczeroshotmonoculardepth}. Two key research directions in this setting centers on diffusion-based completion approaches (e.g., Marigold-DC) \citep{viola2025Marigolddczeroshotmonoculardepth, yu2026large} and direct ViT encoder or decoder adaptation, which refine depth estimates by optimizing a conditioned diffusion model at inference time \citep{seo2026efficienttesttimeoptimizationdepth, ke2026depthcompletionparameterefficienttesttime}.

In contrast to depth completion, metric height completion in the Digital Surface Model (DSM) domain has received limited attention. Existing approaches typically rely on pretrained models that require supervised training on region-specific datasets or sensor-specific data \citep{SONG2026155, chen2026tomosar2height}. We leverage the rich self-supervised feature representations learned by DINOv3, trained on large-scale satellite imagery, which provide high-resolution encoder representations for the first time comparable to natural-image SSL foundation models \citep{siméoni2025dinov3}. These representations provide strong semantic descriptors that can be exploited for geometric calibration at test time. Specifically, we translate these features into local scale and shift parameters that are estimated directly during inference, enabling metric height recovery without domain-specific pretraining.

We further investigate the relationship between feature similarity and height estimation by comparing spatial kNN and semantic kNN strategies \citep{wang2025depthprior, chen2025propagating}. While spatial kNN was widely used in natural-image completion tasks, it assumes that nearby pixels share similar geometric properties. However, in urban environments, repetitive structures such as buildings, roads, and vegetation often exhibit stronger correlations in the feature space than in the spatial domain, particularly when large regions are missing.

Finally, our framework extracts dense features using strided ViT patches. Unlike previous scale-and-shift correction approaches that operate at coarse patch resolution, the strided ViT strategy enables predictions at a much finer spatial resolution \citep{amir2022deepvitfeaturesdense}. This allows the method to produce high-resolution DSM completion directly from backbone features without requiring an additional decoder or task-specific training stage.

\subsection{Monocular model backbone}
Prior2DSM is designed to rescale relative depth maps produced by various relative depth models. It leverages DINOv3 representations together with a lightweight MLP that is optimized during inference. Unlike previous TTA methods that are constrained to specific foundation models \citep{viola2025Marigolddczeroshotmonoculardepth}, Prior2DSM is model-agnostic. Our ablation study shows that MoGe-2 provides better metric proportions compared with Depth Anything V2 (DAV2) and DepthPro, resulting in the best performance. As relative foundation models continue to improve, Prior2DSM is expected to benefit from these advances and achieve even better results.

\subsection{Practical considerations and limitations}

\textit{DSM Updating.}
Prior2DSM can be applied over large DSM-covered regions to update outdated structures. 
To demonstrate this, we used a DSM acquired in 2008 together with a layer indicating regions that have undergone changes. 
The framework generates an updated DSM in these areas, producing a lower reconstruction error compared to the ground-truth updated DSM.

\textit{RGB--DSM Co-Editing.}
Another practical application of Prior2DSM is its integration with generative RS RGB models. 
Here, we demonstrate such an integration using the Text2Earth foundation model for RS text-to-image generation. 
As illustrated in Fig.~\ref{text2earth}, three examples show the generation of new objects together with their corresponding metric DSMs. 
These examples include scenarios such as inserting a single new structure or generating new commercial and residential building regions.

\textit{Limitations.}
Despite its flexibility, Prior2DSM has several limitations. 
First, the method relies on the quality of the relative depth input, and errors in the initial depth estimation may propagate to the final metric DSM reconstruction. 
Second, the TTA optimization process introduces additional computational cost. 
In our implementation, the optimization process requires approximately 8.2 minutes per square kilometer for 100 optimization steps on an RTX~4000 GPU. In practical scenarios involving individual buildings or small neighborhoods, the framework can update the DSM within a few seconds. While this computational cost is acceptable for localized updates, further improvements in efficiency would be beneficial for large-scale deployment.

\section{Conclusions}
We propose a TTA framework built upon strong foundation models for metric DSM completion. Our comprehensive evaluation demonstrates that Prior2DSM outperforms pretrained depth completion backbone networks across different levels of incompleteness on both aerial and satellite imagery. First, to the best of our knowledge, this is the first model designed for zero-shot height completion using a frozen DINOv3 satellite foundation model. Second, Prior2DSM employs TTA LoRA and MLP adaptations to further adjust the attention responses and estimate scale and shift parameters for each scene. This design mitigates biases toward domain-specific training and enables the framework to operate with any relative depth prediction. Third, we provide an ablation study comparing multiple relative depth models to analyze their influence on the final metric DSM reconstruction. We hope this work will advance the practical deployment of height completion models and inspire future research in this direction.

\section*{Acknowledgment}

We thank the Ministry of Agriculture, Chief Scientist Program, grant number 16-17-0005, 2022, and the Negev Scholarship from the Kreitman School of Ben-Gurion University of the Negev for supporting Osher Rafaeli’s PhD studies.

\bibliographystyle{elsarticle-harv}
\bibliography{references}

\end{document}